\documentclass[10pt,twocolumn,letterpaper]{article}

\usepackage[pagenumbers]{preprint}

\usepackage[utf8]{inputenc}
\usepackage[T1]{fontenc}
\usepackage{times}
\usepackage{epsfig}
\usepackage{graphicx}
\usepackage{amsmath}
\usepackage{amssymb}
\usepackage{amsthm}
\usepackage{multirow}
\usepackage{booktabs}
\usepackage{colortbl}
\usepackage{xcolor}
\usepackage{algorithm}
\usepackage{algpseudocode}
\usepackage{subcaption}

\usepackage[numbers,sort&compress]{natbib}

\definecolor{linkblue}{rgb}{0.21,0.49,0.74}
\usepackage[pagebackref,breaklinks,colorlinks,allcolors=linkblue]{hyperref}

\definecolor{selectedrow}{RGB}{211,228,251}

\title{Geometry-Aware Implicit Memory for Video World Models}

\author{%
Zhengxuan Wei$^{1,2,*}$ \quad
Xu Guo$^{3,2,*}$ \quad
Xinghui Li$^{2,*,\dagger}$ \quad
Xunzhi Xiang$^{1}$ \quad
Min Wei$^{2}$ \quad
Yiran Zhu$^{2}$ \\
Qiulin Wang$^{2}$ \quad
Xintao Wang$^{2}$ \quad
Pengfei Wan$^{2}$ \quad
Xiangwang Hou$^{3}$ \quad
Qi Fan$^{1,\dagger}$
\\[4pt]
$^{1}$School of Intelligence Science and Technology, Nanjing University \\
$^{2}$Kling Team, Kuaishou Technology \quad
$^{3}$Tsinghua University
\\[4pt]
{\small Project page: \url{https://gim-world.github.io/}}
}

\makeatletter
\let\@oldmaketitle\@maketitle
\renewcommand{\@maketitle}{%
  \@oldmaketitle
  \vspace{-30pt}
  \begin{center}
    \includegraphics[width=\textwidth]{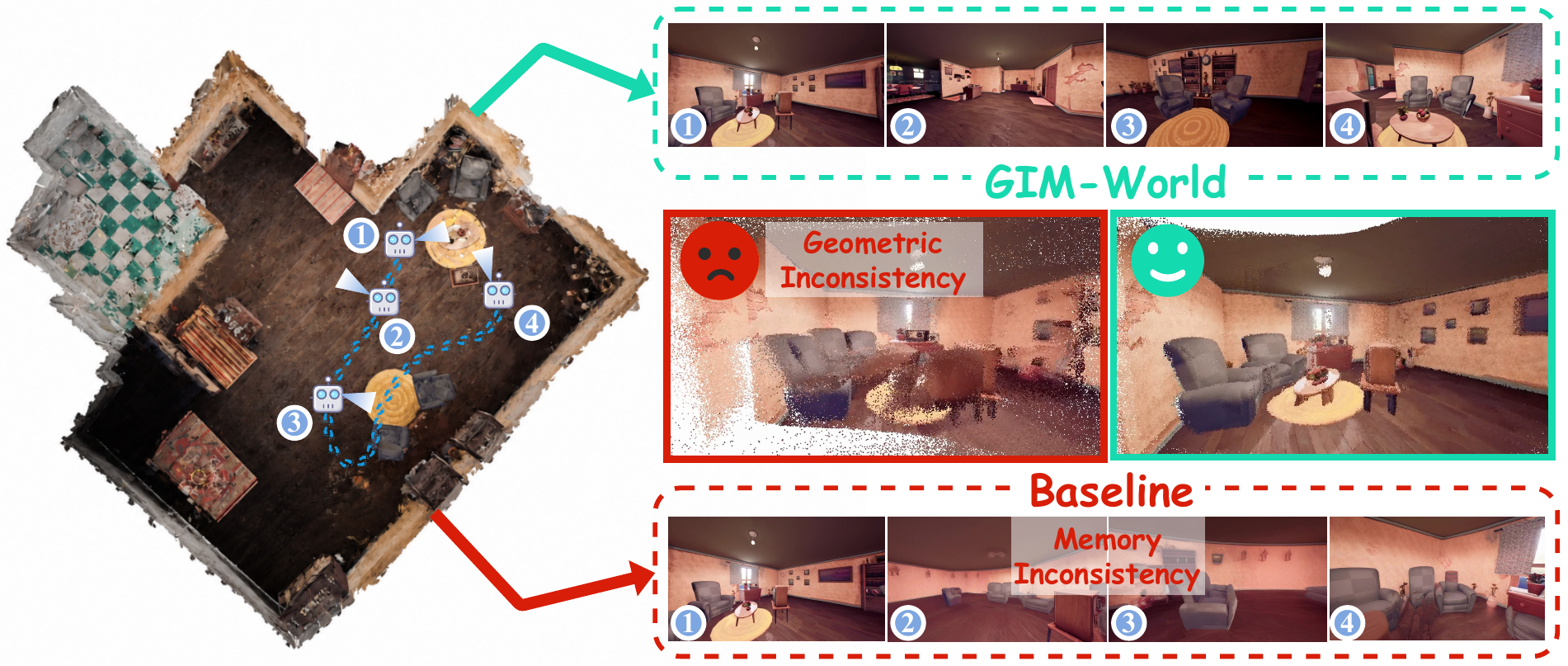}\\[0.5em]
    \begin{minipage}{\textwidth}
      \captionof{figure}{We present \textbf{GIM-World}, a geometry-aware
      implicit memory framework for video world models. By distilling 3D
      scene structure from a frozen 3D foundation model into a compact,
      fixed-size memory state, GIM-World produces long-horizon
      autoregressive rollouts that remain geometrically coherent and
      visually consistent with the underlying scene, while existing
      memory designs degrade into visible geometric distortion and
      memory drift.}
      \label{fig:teaser}
    \end{minipage}
  \end{center}
  \vspace{-5pt}
}
\makeatother

\begin{document}
\maketitle

{\let\thefootnote\relax%
  \footnotetext{$^{*}$Equal contribution.}%
  \footnotetext{$^{\dagger}$Corresponding author.}}

\begin{abstract}
Video world models aim to simulate controllable visual environments, but
long-horizon rollouts depend on what the model remembers after observations
leave its native context window. Explicit memories retain frames or online
3D reconstructions, which can suffer from heuristic retrieval errors,
redundant appearance storage, or reconstruction artifacts. Implicit memories
compress history into a compact state, but existing designs are not explicitly
constrained to encode cross-view scene geometry. We propose
\textbf{GIM-World}, a geometry-aware implicit memory framework for video
world models. A lightweight transformer encoder compresses variable-length
history into fixed-size memory tokens, a camera-queryable geometry head
distills 3D scene structure from a frozen foundation model into the memory
during training, and an information-guided pruning rule keeps encoding cost
bounded as history grows. The geometry teacher is discarded at inference,
leaving a lightweight memory module. Experiments on MIND show that
GIM-World better preserves long-horizon geometric and visual consistency
than both explicit- and implicit-memory baselines.
\end{abstract}

\section{Introduction}

Video world models learn controllable visual simulators. Given past
observations together with a stream of actions or camera motion, they
predict how the scene evolves next~\citep{ha2018world,bruce2024genie,
decart2024oasis,sun2025worldplay,he2025matrix}. As rollouts grow
beyond a few seconds, per-frame realism alone is no longer sufficient,
and long-horizon consistency becomes the central challenge. Regions
an agent has visited before should look the same when it returns to
them, forward and reverse trajectories should agree on scene layout,
and newly generated viewpoints should remain compatible with structure
established earlier. These properties decide whether the rollout can
support embodied agents, interactive games, and persistent visual
simulators~\citep{team2026advancing,duan2026liveworld,russell2025gaia,
ren2025cosmos}. They are decided, in turn, by what the model keeps
about observations that have left its native context window.

Existing memory mechanisms fall into two families. Explicit memory keeps
past observations as explicit items, either historical frames that the
generator retrieves at each rollout step~\citep{yu2025context,
xiao2025worldmem,oshima2025worldpack,chen2026out,hong2025relic},
or online 3D reconstructions of the scene~\citep{ren2025gen3c,li2025vmem,
wu2025video,xu2026ucm}. Frame-level retrieval typically relies on
heuristic rules such as visual similarity or camera overlap, and a
missed or wrong selection can propagate as an inconsistency in the
generated rollout, while appearance redundancy among retained frames
further limits how much scene information any bounded memory can
actually carry. 3D reconstructions avoid this heuristic selection,
but they require running a reconstruction pipeline alongside the
generator, adding inference latency and engineering complexity, and
can be sensitive to dynamic objects, occlusion, or weak texture, with
reconstruction errors that can accumulate over long rollouts. Implicit memory instead uses a memory model decoupled from
the generator, compressing past frames into a compact world-state
representation~\citep{wu2025pack,savov2025statespacediffuser,
wu2026infinite}. Current implicit memories, however, lack a geometric
training signal. The appearance signals they compress are highly
redundant across views, and the resulting compressed state is not
explicitly constrained to encode cross-view scene structure.

Meanwhile, 3D foundation models~\citep{wang2024dust3r,leroy2024grounding,
wang20253d,wang2025continuous,wang2025vggt} recover dense per-frame
geometric features from video in a single feed-forward pass. A growing
line of work injects these features into video generators along three
axes: direct conditioning, which concatenates geometric tokens or
point-map projections to the generator's
input~\citep{zhang2025world,cao2025uni3c,ren2025gen3c,hu2026geometry,
huang2026cinescene,xie2026lavr}; feature supervision, which aligns the
generator's backbone features with 3D features at training
time~\citep{wu2025geometry,yu2024representation}; and post-training alignment,
which derives rewards or preferences from 3D features in an additional DPO
or RL stage~\citep{du2026videogpa,kupyn2025epipolar,an2026vggrpo}.
Despite their differences, these methods primarily inject geometry
into the generator, either as input, supervision, or post-training
feedback. We take the opposite view. Geometry is best treated as a
property of the remembered world rather than only as an additional
signal on the generator; it should therefore be encoded in the memory
state.

We propose \textbf{GIM-World}, a geometry-aware implicit memory framework
for video world models. It has three parts, introduced in turn below: a
memory encoder, a training-only geometry supervision head, and an
information-guided pruning rule. The memory itself is formed by a simple
transformer-based encoder: a fixed-size set of learnable memory queries
attends to the tokens of the historical frames, and the updated queries
are concatenated with the target latents and passed to the generator as
conditioning. Memory size therefore does not grow with the history. The
encoder is also lightweight at inference, running in less than $0.3\%$
of the diffusion backbone's time, so the fixed-size memory comes at
negligible compute cost.

To force the memory to carry geometry, we attach a lightweight
camera-queryable geometry head. Rather than aligning generator
features with 3D features or reconstructing historical frames
directly, we train the memory to answer camera-conditioned geometric
queries. Given the memory and a historical camera, the head predicts
the per-patch features that a 3D foundation model would extract from
that view. The head is trained jointly with the memory encoder and
the generator, while the 3D foundation model is kept frozen and used
only as a training-time teacher. All trainable components share a
single end-to-end objective, so the framework is trained in one stage
without staged schedules or auxiliary fine-tuning. At inference, the
geometry head and the 3D teacher are discarded.

The memory has a fixed size, but the cost of building it still scales
with the number of history tokens. We prune history before encoding by
keeping the subset that best predicts the frames we drop, following the
mutual-information sensor-placement criterion of~\citet{krause2008near},
and optimize it with a greedy algorithm.

Our contributions are as follows:
\begin{itemize}
    \item We propose a lightweight implicit memory framework for video
    world models. A memory encoder compresses variable-length history
    into a fixed-size set of memory tokens, and an information-guided
    pruning rule keeps its cost bounded as the history grows. 
    \item We design a camera-queryable geometry supervision loss that
    distills scene geometry from a frozen 3D foundation model into the
    memory during training, so the compressed state encodes cross-view
    scene structure rather than just appearance.
    \item Experiments show that GIM-World consistently outperforms both
    explicit-memory and implicit-memory baselines on memory consistency,
    action control, and 3D geometric consistency.
\end{itemize}

\section{Related Work}

\subsection{Video World Models}

Classical world models compress interaction histories into latent dynamics for
planning and control \citep{ha2018world,hafner2023mastering}. Recent
video world models instead learn controllable visual simulators directly from
interaction data, predicting future observations in pixel or latent video
space given past observations together with actions, cameras, or user
controls. This direction has developed rapidly in games and simulated
environments \citep{bruce2024genie,alonso2024diffusion,valevski2024diffusion,
decart2024oasis,che2024gamegen,feng2024matrix,yu2025gamefactory,
guo2025mineworld,huang2025vid2world}, with more recent interactive generators
emphasizing real-time or streaming control with persistent scene consistency
\citep{mao2025yume,he2025matrix,sun2025worldplay,team2026advancing}.

Recent work has increasingly moved beyond short reactive prediction toward
persistent and interactive world simulation, with systems exploring real-time
interaction, out-of-sight dynamics, 4D scene simulation, and multi-agent or
multi-view settings \citep{team2026advancing,duan2026liveworld,
team2026inspatio,chen2025teleworld,wu2026multiworld}. Similar generative
world-model ideas also appear in driving and robotics
\citep{russell2025gaia,ren2025cosmos,zhu2025unified,guo2025ctrl}. These
efforts motivate our focus on the memory needed to support persistent world
simulation.

\subsection{Memory in Video Generation and World Models}

Long videos are often generated by extending the rollout or the available
context, using denoising windows, diffusion or self-forcing, and routed context
\citep{qiu2023freenoise,kim2024fifo,chen2024diffusion,
huang2025self,cai2025mixture,xiang2025make,xiang2025macro,
xiang2026pathwise}. These methods increase the
generation horizon, but they leave open what information should remain once
observations fall outside the model's native context window.

Memory-based systems address this by keeping history in different forms.
Retrieval and packing methods keep frames, compressed tokens, or KV caches for
later reuse
\citep{yu2025context,xiao2025worldmem,gao2026memcam,
guo2026memorize,oshima2025worldpack,chen2026out,hong2025relic}.
Spatial memories bind history to explicit geometric substrates such as 3D caches,
surfels, point maps, or warped positional structures
\citep{ren2025gen3c,li2025vmem,wu2025video,huang2025memory,
yu2026mosaicmem,xu2026ucm}. More recent implicit memories learn compact states
whose size does not grow with the full history
\citep{wu2025pack,savov2025statespacediffuser,yu2025videossm,
wu2026infinite}. Our work
follows the implicit-memory direction but differs in how the compressed
state is trained. Rather than relying only on the future-frame generation
loss, we supervise the memory to answer camera-conditioned geometric
queries, so that it encodes cross-view scene structure.

\subsection{Geometry Consistency in Video Generation}

Camera-controlled video generation exposes a central weakness of purely
image-space training: visually plausible rollouts can still drift in camera
motion, parallax, or scene structure. One line of work supplies geometry to the
generator, from camera and motion control to RGB-XYZ modeling, point-cloud or
3D-cache conditioning, geometry-as-context, and implicit 3D/4D feature
conditions \citep{he2024cameractrl,wang2024motionctrl,zhang2025world,
cao2025uni3c,ren2025gen3c,hu2026geometry,huang2026cinescene,
xie2026lavr}. Another line uses geometry as an optimization signal:
Geometry Forcing aligns video features with VGGT representations, while
VideoGPA, Epipolar-DPO, and VGGRPO derive preference or reward signals from
geometric consistency \citep{wu2025geometry,du2026videogpa,
kupyn2025epipolar,an2026vggrpo}. These methods improve the geometric
fidelity of generated videos, but the geometric signal is applied to the
generator or its outputs. Our focus is to use such supervision to shape the
fixed-size memory state itself, with no geometry model or head at inference.

\section{Methodology}

\begin{figure*}[t]
  \centering
  \includegraphics[width=0.95\textwidth]{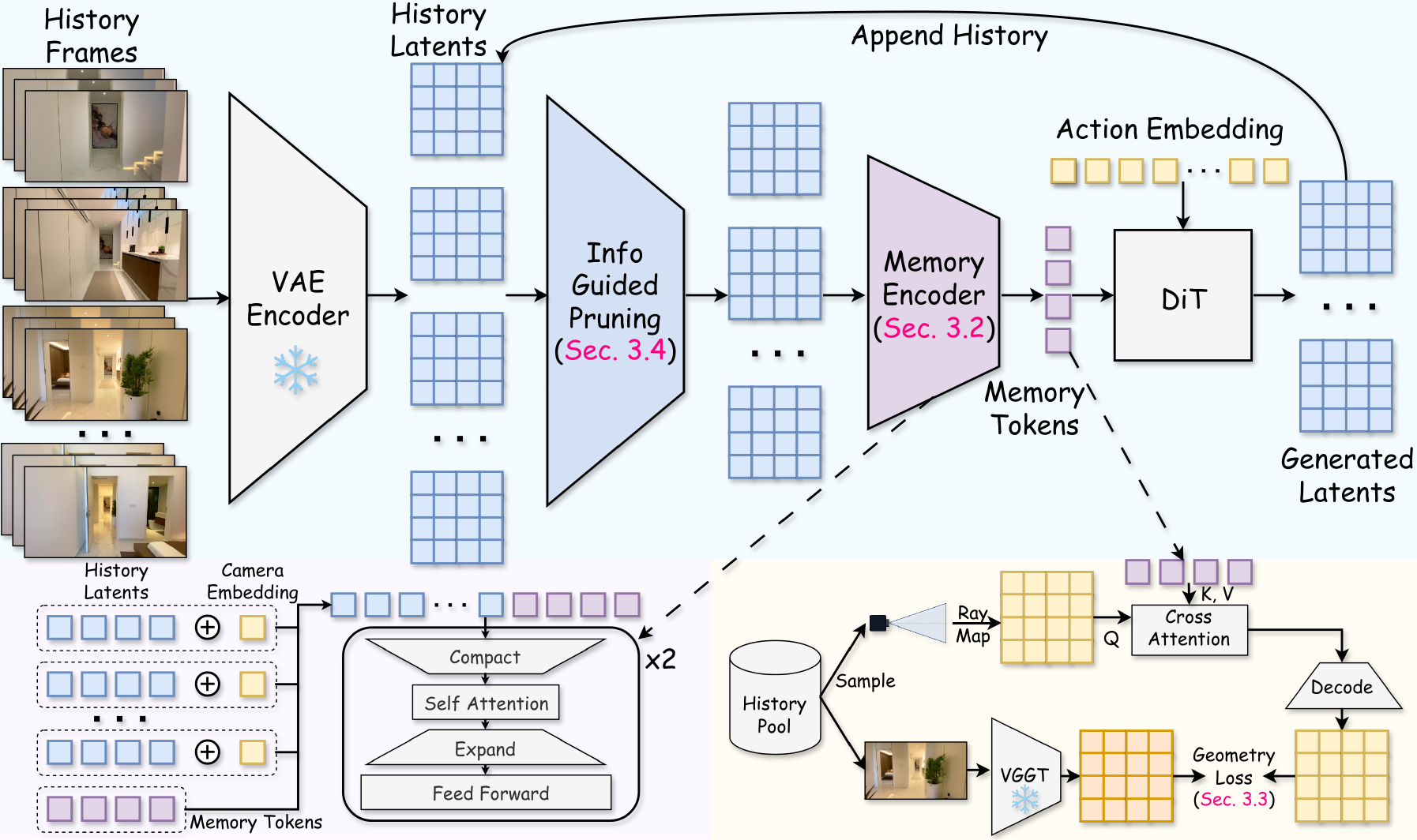}
  \caption{\textbf{Overview of GIM-World.}
    \emph{Top}: history latents are pruned (Sec.~\ref{sec:method_prune}) and
    compressed into a small set of memory tokens
    (Sec.~\ref{sec:method_arch}), which condition the DiT together with
    action embeddings to generate the next latents.
    \emph{Bottom-left}: the memory encoder fuses history latents with
    camera embeddings and the memory tokens through two blocks of compact
    self-attention and feed-forward layers.
    \emph{Bottom-right}: during training, a sampled frame's camera produces
    a ray map that queries the memory; the decoded output is matched against
    its VGGT features through a geometry loss
    (Sec.~\ref{sec:method_geo}).}
  \label{fig:framework}
\end{figure*}

We introduce a geometry-aware implicit memory that writes past observations
and their camera poses into a fixed-size state, and supervises that state to
reconstruct the geometry features corresponding to a given viewpoint.
Fig.~\ref{fig:framework} gives an overview of the framework.
Sec.~\ref{sec:method_problem} formulates the memory-conditioned world-modeling
problem, and Sec.~\ref{sec:method_arch} instantiates the memory as a fixed-size set of
tokens produced by a learned encoder. To prevent this encoder from
merely caching appearance, Sec.~\ref{sec:method_geo} introduces
camera-queryable geometry supervision. For long histories, Sec.~\ref{sec:method_prune}
additionally prunes redundant observations under an information-guided
criterion. Finally, Sec.~\ref{sec:method_train} summarizes the training
objective and the inference pipeline.

\subsection{Problem Formulation}
\label{sec:method_problem}

A video world model predicts a future rollout by iteratively generating new
frames from past observations and future actions. At rollout step $t$, given
the observed history
\begin{equation}
    \mathcal{H}_t=\{(x_i,c_i)\}_{i=1}^{t},
\end{equation}
where $x_i$ is an observed frame (or its VAE latent) and $c_i$ is its camera
pose, and a block of future actions $a_{t+1:t+K}$, the model samples the next
$K$ frames,
\begin{equation}
    x_{t+1:t+K}\sim p_\theta(\cdot \mid \mathcal{H}_t, a_{t+1:t+K}),
\end{equation}
and appends them, together with their cameras, to form $\mathcal{H}_{t+K}$.
The process repeats until the full rollout is complete.

Conditioning on the unbounded $\mathcal{H}_t$ does not scale with rollout
length. Memory-based world models instead summarize the history through a
memory operator $\mathcal{M}$ and the corresponding memory state
\begin{equation}
    m_t = \mathcal{M}(\mathcal{H}_t),
\end{equation}
and generate from
\begin{equation}
    p_\theta(x_{t+1:t+K}\mid m_t, a_{t+1:t+K}).
\end{equation}
The form of $\mathcal{M}$ defines the design space of memory-based world
models: it can retrieve frames, cache tokens, build an explicit 3D
reconstruction, or encode a learned implicit state.

\subsection{Implicit Memory Encoder}
\label{sec:method_arch}

The memory encoder maps a variable-length, camera-indexed history to a fixed
number of memory slots. Historical frames are first turned into patch tokens
by the video VAE and the diffusion backbone's patch embedding, yielding
\begin{equation}
    H=\{h_{i,p}\}_{i=1,p=1}^{T,P},
\end{equation}
where each frame lives on a patch grid of $H_p\times W_p$ with
$P=H_p W_p$ tokens. Since the same scene point can appear from many
viewpoints, we inject camera information only into the history tokens,
\begin{equation}
    \tilde{h}_{i,p}=h_{i,p}+E_c(c_i),
\end{equation}
where $E_c$ is a linear camera embedding. The memory queries themselves stay
pose-free and are therefore not tied to a particular view or future action.

We maintain $N_m$ learnable memory queries $Q_0\in\mathbb{R}^{N_m\times d}$,
and concatenate them with the camera-aware history into the encoder input
$Z_0=[Q_0;\tilde{H}]$. The encoder uses only two blocks, which keeps it
lightweight and adds negligible cost to the diffusion backbone at inference
time. Each block updates $Z$ through a compact self-attention branch and a
full-resolution feed-forward branch:
\begin{align}
    Z &\leftarrow Z + \mathrm{Expand}\big(\mathrm{Attn}(\mathrm{Compact}(Z))\big), \\
    Z &\leftarrow Z + \mathrm{FFN}(Z).
\end{align}

The two operators act per frame:
\begin{align}
    \mathrm{Compact}&: \mathbb{R}^{H_p\times W_p\times d}
        \to \mathbb{R}^{(H_p/s)\times (W_p/s)\times d}, \\
    \mathrm{Expand}&: \mathbb{R}^{(H_p/s)\times (W_p/s)\times d}
        \to \mathbb{R}^{H_p\times W_p\times d},
\end{align}
where $\mathrm{Compact}$ flattens each $s\times s$ block of tokens through a
shared linear layer applied identically to queries and history, and
$\mathrm{Expand}$ inverts this with another shared linear layer.
Self-attention therefore runs on $(N_m+T\,H_p W_p)/s^2$ tokens, with separate
rotary positional grids for the query and history segments.

After the two blocks, we take $m$ as the first $N_m$ tokens of $Z$, and feed
$m$ to the diffusion backbone by concatenating it along the temporal axis
with the target latents, following Context-as-Memory
\citep{yu2025context}.

\subsection{Camera-Queryable Geometry Supervision}
\label{sec:method_geo}

Flow matching trains the generator to predict future frames, but it does not
specify what information the memory should store. A fixed-size implicit memory
could spend capacity on redundant appearance projections that are useful for
short-term reconstruction but poor as a compact world state. We therefore add a
training-time geometry supervision signal that directly constrains the memory.

Recent methods such as Geometry Forcing~\citep{wu2025geometry} inject
geometric information into video generators by aligning the intermediate
features of the backbone with per-frame encoder features from a 3D foundation
model such as VGGT~\citep{wang2025vggt}. This assumes the student
representation stays indexed by frame and spatial location, so that alignment
can be applied token-to-token. Our memory is deliberately different: after
compressing a variable-length history, its tokens carry no one-to-one
correspondence with input frames or patches, and direct token matching would
impose an artificial ordering on the memory slots and collapse the
representation back toward a frame-indexed cache. Instead, we query the
memory through historical cameras and align the queried output against VGGT
features.

Concretely, at each training step we uniformly sample one historical frame
$i$ from $\mathcal{H}_t$. Conditioned on the sampled camera $c_i$, a
lightweight geometry head maps the memory $m$ to a feature map
$\hat{G}_i\in\mathbb{R}^{H_gW_g\times d_g}$ defined on an $H_g\times W_g$
patch grid, which we supervise against the VGGT encoder feature map
$G_i\in\mathbb{R}^{H_gW_g\times d_g}$ of the same frame.

To condition the geometry head on the queried viewpoint, we convert $c_i$ into a
set of per-patch queries. For patch $(u,v)$, the world-space ray
\begin{equation}
    \rho(c_i)_{u,v}=[o_i,d_{i,u,v}]\in\mathbb{R}^{6}
\end{equation}
concatenates the camera origin $o_i$ and the normalized ray direction
$d_{i,u,v}$; it is embedded by a small MLP and added to a learnable 2D grid
positional embedding $e_{u,v}$ shared across frames,
\begin{equation}
    q_{i,u,v}=\mathrm{MLP}_{ray}(\rho(c_i)_{u,v})+e_{u,v}.
\end{equation}
The queries $q_i$ then pass through the geometry head: a
cross-attention over the memory, a self-attention over the patch grid, and
a feed-forward network that projects the output into the VGGT encoder
feature dimension,
\begin{equation}
    \hat{G}_i = \mathrm{FFN}\bigl(\mathrm{SelfAttn}(\mathrm{CrossAttn}(q_i,\,m))\bigr).
\end{equation}
The cross-attention retrieves the information from $m$ that answers the ray
query, while the self-attention exchanges context across the patch grid.

We train the memory and the geometry head jointly with a per-patch cosine loss
between $\hat{G}_i$ and $G_i$,
\begin{equation}
    \mathcal{L}_{geo}
    =
    1-\frac{1}{H_gW_g}
    \sum_{u,v}
    \cos\bigl(\hat{G}_{i,u,v},\,G_{i,u,v}\bigr).
\end{equation}
Since $i$ is resampled at every step, this single-frame estimator is
unbiased for the all-frame alignment objective, while the per-step
distillation cost stays independent of the history length.

\subsection{Information-Guided Pruning}
\label{sec:method_prune}

The memory has fixed size, but the cost of forming it still scales with
the number of history tokens. For long rollouts, many views are redundant:
nearby cameras often observe nearly the same scene through slightly
different projections. We prune history before memory encoding by keeping
a retained subset $S\subseteq\mathcal{H}$ with $|S|\le K$ that is as
informative as possible about the frames it leaves out, following the
mutual-information sensor-placement criterion
of~\citet{krause2008near}:
\begin{equation}
    S^\star
    =
    \arg\max_{S}\; I\bigl(S;\,\mathcal{H}\setminus S\bigr)
    \quad\text{s.t.}\quad |S|\le K.
\end{equation}

To evaluate $I(S;\mathcal{H}\setminus S)$, we place a Gaussian process
over per-frame observations with a pose-time kernel
\begin{equation}
\label{eq:pose_time_kernel}
    k(c_i,c_j)
    =
    \exp\!\left(
        -\frac{\lVert p_i-p_j\rVert_2^2}{2\sigma_p^2}
        -\frac{\angle(\mathbf{f}_i,\mathbf{f}_j)^2}{2\sigma_r^2}
        -\frac{(t_i-t_j)^2}{2\sigma_t^2}
    \right),
\end{equation}
where $p_i$ is the camera position, $\mathbf{f}_i$ the camera forward
direction, $t_i$ the frame time, and $\sigma_{p},\sigma_{r},\sigma_{t}$
are RBF bandwidths. Two frames are highly correlated under this GP when
their cameras are close in space, point in similar directions, and are
close in time, so frames likely to observe similar scene content are
treated as similar.

The GP turns the kernel into a posterior variance $\sigma^2(h\mid A)$
that measures the remaining uncertainty about frame $h$ after observing
frames in $A\subseteq\mathcal{H}$:
\begin{equation}
    \sigma^2(h\mid A) = k(h,h) - \mathbf{k}_{hA}^\top K_{AA}^{-1} \mathbf{k}_{Ah},
\end{equation}
where $K_{AA}=[k(a_i,a_j)]$ is the kernel matrix over $A$ and
$\mathbf{k}_{hA}=[k(h,a_i)]$ the kernel vector between $h$ and the frames
in $A$. The first term $k(h,h)$ is the prior uncertainty about $h$; the
second term subtracts the portion explained by observing $A$. So
$\sigma^2(h\mid A)$ is large when $h$ has low kernel similarity to every
frame in $A$, and small when $h$ is close to at least one frame of $A$.

$I(S;\mathcal{H}\setminus S)$ is submodular, and the standard greedy
rule is near-optimal for this objective~\citep{krause2008near}. At step
$t$, let $\bar{S}_{t-1}=\mathcal{H}\setminus S_{t-1}$ denote the
still-unselected frames. Under the GP, the greedy marginal gain for
adding $h\in\bar{S}_{t-1}$ is a log-ratio of posterior variances,
\begin{equation}
    s_t
    =
    \arg\max_{h\in\bar{S}_{t-1}}\;
    \tfrac{1}{2}\log\frac{\sigma^2\!\bigl(h \mid S_{t-1}\bigr)}
             {\sigma^2\!\bigl(h \mid \bar{S}_{t-1}\setminus\{h\}\bigr)}.
\end{equation}
The numerator is large when the retained set $S_{t-1}$ cannot predict
$h$, so adding $h$ brings new information into $S$; the denominator is
small when the other unselected frames already predict $h$, so $h$ is
representative rather than an outlier. Greedy therefore fills $S$ with
frames that are informative relative to the current retained set yet
well-covered by the rest of the history.

\subsection{Training and Inference}
\label{sec:method_train}

The diffusion backbone is trained with a flow-matching objective conditioned
on the memory and the future action sequence,
\begin{equation}
    \mathcal{L}_{FM}
    =
    \mathbb{E}\bigl[\|v_\theta(z_t,t,m,a)-v^\star\|_2^2\bigr],
\end{equation}
and the memory encoder, the geometry head, and the backbone are optimized jointly
to minimize
\begin{equation}
    \mathcal{L} = \mathcal{L}_{FM} + \lambda\,\mathcal{L}_{geo}.
\end{equation}
The whole pipeline is therefore trained end-to-end in a single
stage, without separate pretraining, distillation, or fine-tuning
steps. At inference, the geometry head and the VGGT encoder are no
longer needed and are discarded.

\section{Experiments}

\begin{table*}[t]
  \centering
  \small
  \setlength{\tabcolsep}{3pt}
  \setlength{\abovecaptionskip}{2pt}
  \setlength{\belowcaptionskip}{2pt}
  \renewcommand{\arraystretch}{1.05}
  \caption{
    Quantitative comparison on the first-person and third-person MIND
    test sets. Memory metrics quantify how faithfully each method
    reconstructs the observed scene, action accuracy reports the relative
    pose error of the rendered camera trajectory, and 3D Geometry reports
    the normalized dense reprojection score. Lower is better for MSE,
    LPIPS, and RPE; higher is better for PSNR, SSIM, and the reprojection
    score.
  }
  \label{tab:mind_main_results}
  \resizebox{\textwidth}{!}{%
  \begin{tabular}{lccccccc ccccccc}
    \toprule
    \multirow{2.5}{*}{Method}
    & \multicolumn{7}{c}{First-person view}
    & \multicolumn{7}{c}{Third-person view} \\
    \cmidrule(lr){2-8}
    \cmidrule(lr){9-15}
    & \multicolumn{4}{c}{Memory}
    & \multicolumn{2}{c}{Action Accuracy}
    & \multicolumn{1}{c}{3D Geometry}
    & \multicolumn{4}{c}{Memory}
    & \multicolumn{2}{c}{Action Accuracy}
    & \multicolumn{1}{c}{3D Geometry} \\
    \cmidrule(lr){2-5}
    \cmidrule(lr){6-7}
    \cmidrule(lr){8-8}
    \cmidrule(lr){9-12}
    \cmidrule(lr){13-14}
    \cmidrule(lr){15-15}
    & MSE$\downarrow$ & PSNR$\uparrow$ & SSIM$\uparrow$ & LPIPS$\downarrow$
    & Trans.$\downarrow$ & Rot.$\downarrow$ & Reproj.$\uparrow$
    & MSE$\downarrow$ & PSNR$\uparrow$ & SSIM$\uparrow$ & LPIPS$\downarrow$
    & Trans.$\downarrow$ & Rot.$\downarrow$ & Reproj.$\uparrow$ \\
    \midrule
    \multicolumn{15}{c}{\textit{Explicit Memory}} \\
    \midrule
    Matrix-Game 2.0
    & 0.1188 & -- & -- & -- & 0.0265 & 0.6914 & --
    & 0.1404 & -- & -- & -- & 0.0622 & 0.9031 & -- \\
    MIND-World
    & 0.1035 & -- & -- & -- & 0.0384 & \underline{0.5534} & --
    & 0.1042 & -- & -- & -- & 0.0321 & 0.3328 & -- \\
    FramePack
    & 0.0764 & 12.04 & 0.3763 & 0.7062 & 0.0269 & 0.6536 & \underline{73.97}
    & 0.0735 & 12.23 & 0.3957 & 0.6770
    & \underline{0.0145} & \underline{0.2852} & \underline{70.82} \\
    Context-as-Memory
    & \underline{0.0706} & \underline{12.50} & 0.3917 & \underline{0.6953}
    & \textbf{0.0235} & 0.5697 & 66.14
    & \underline{0.0671} & \underline{12.81} & \underline{0.4051} & \underline{0.6480}
    & 0.0241 & 0.4385 & 69.17 \\
    \midrule
    \multicolumn{15}{c}{\textit{Implicit Memory}} \\
    \midrule
    SSM
    & 0.0796 & 11.96 & \underline{0.3953} & 0.7439 & 0.0379 & 0.6702 & 53.95
    & 0.0928 & 11.76 & 0.4037 & 0.6830 & 0.0201 & 0.2971 & 61.26 \\
    GIM-World
    & \textbf{0.0614} & \textbf{13.40} & \textbf{0.4135} & \textbf{0.6304}
    & \underline{0.0247} & \textbf{0.4126} & \textbf{81.70}
    & \textbf{0.0605} & \textbf{13.58} & \textbf{0.4303} & \textbf{0.5974}
    & \textbf{0.0106} & \textbf{0.1588} & \textbf{87.10} \\
    \bottomrule[1pt]
  \end{tabular}%
  }
\end{table*}

\begin{figure*}[t]
  \centering
  \includegraphics[width=\linewidth]{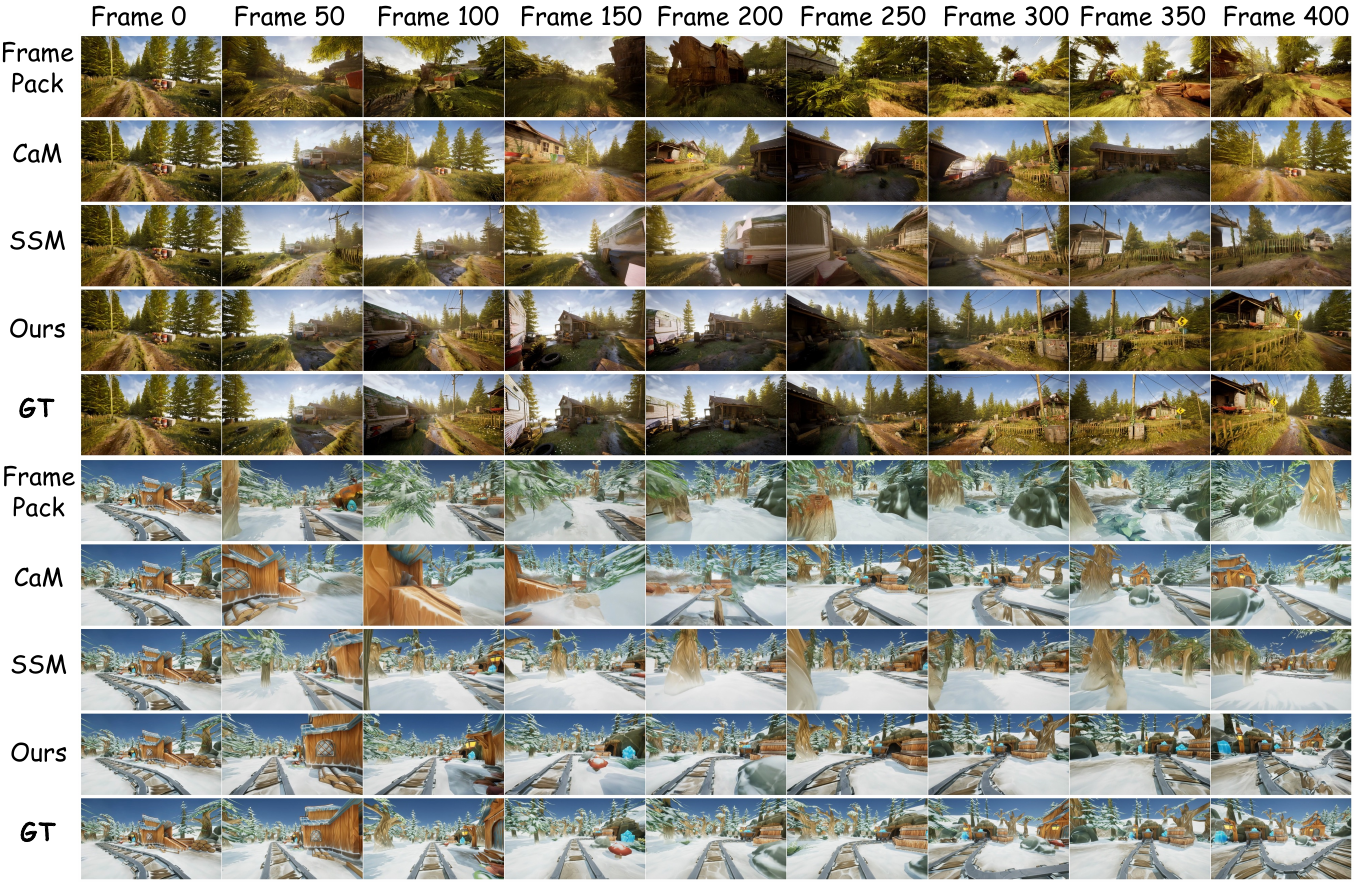}
  \caption{Qualitative comparison on first-person MIND scenes.}
  \label{fig:qual_1st}
\end{figure*}

\begin{figure*}[t]
  \centering
  \includegraphics[width=\linewidth]{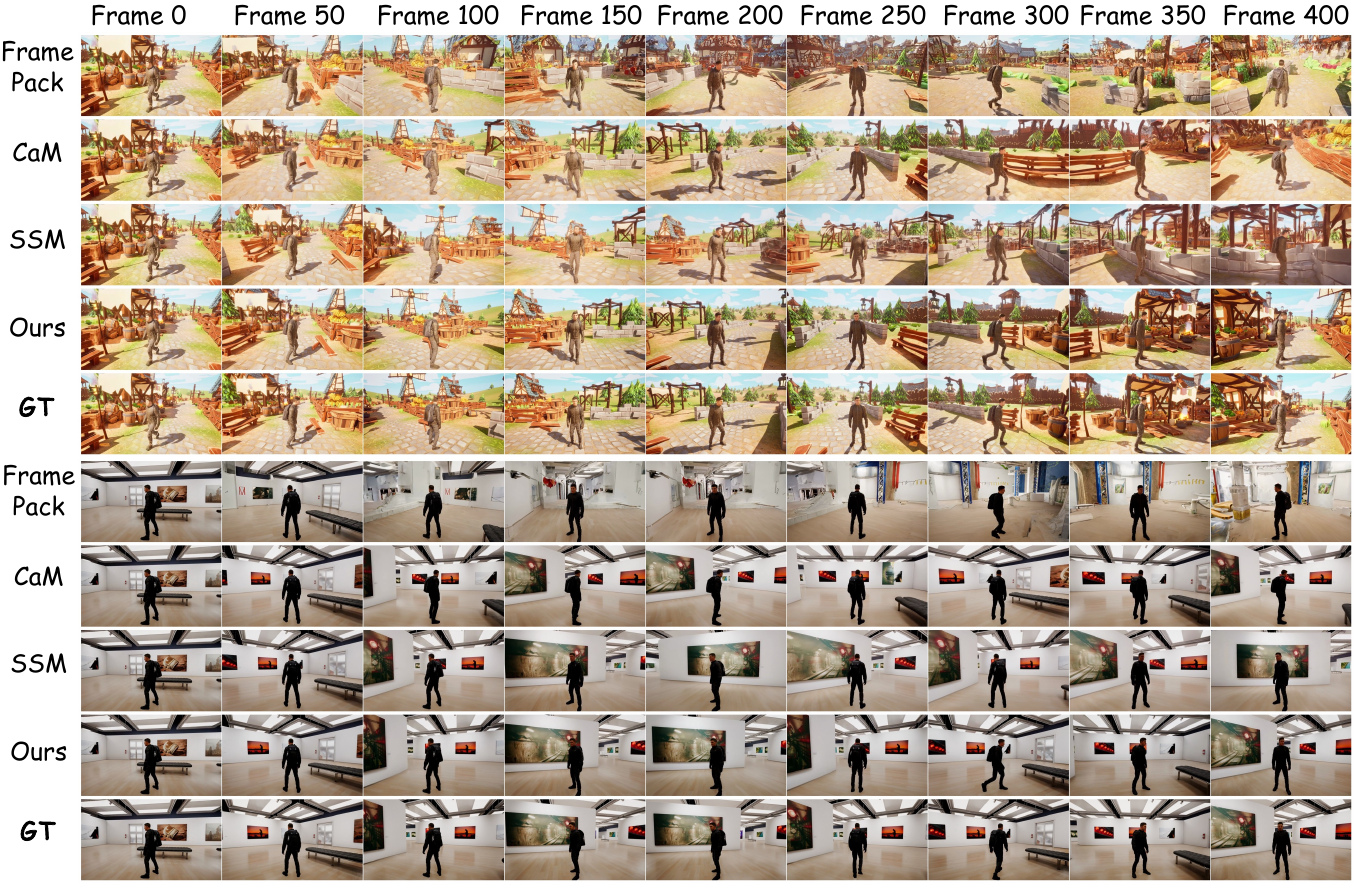}
  \caption{Qualitative comparison on third-person MIND scenes.}
  \label{fig:qual_3rd}
\end{figure*}

\begin{figure*}[t]
  \centering
  \includegraphics[width=\linewidth]{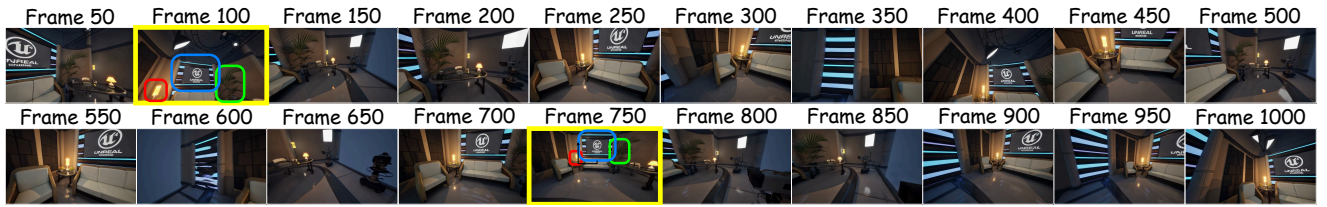}
  \caption{Long-horizon qualitative results on the MIND benchmark.}
  \label{fig:long_horizon}
\end{figure*}

\subsection{Experimental Setup}
\label{sec:exp_setup}

\paragraph{Dataset.}
We train and evaluate on MIND~\citep{ye2026mind}, an open-domain
closed-loop revisited benchmark for video world models. MIND provides
$1080$p / $24$\,fps Unreal Engine 5 videos with a shared action space
of four planar movement keys $W,A,S,D$ and four camera rotation keys
$\uparrow,\downarrow,\leftarrow,\rightarrow$. The benchmark contains
$100$ first-person and $100$ third-person clips, each about three
minutes long, split $50/50$ into training and evaluation; we follow
this split exactly. Each evaluation clip is annotated with a context
boundary that takes roughly the first quarter of the clip as context
and leaves the remaining three quarters as the prediction segment.
Following MIND's protocol, the model observes the context as initial
memory and then autoregressively generates the prediction segment
under the same action sequence; evaluation is performed entirely on
the prediction segment, at $720$p resolution, separately for the
first-person and third-person subsets.

\paragraph{Baselines.}
We group baselines along the explicit-versus-implicit memory axis used
in the introduction. For \emph{explicit memory}, we compare against
Matrix-Game 2.0~\citep{he2025matrix} and
MIND-World~\citep{ye2026mind}, taking their numbers directly from the
MIND benchmark, and against FramePack~\citep{zhang2025framepack} and
Context-as-Memory~\citep{yu2025context}, which we re-train under our
setting. For \emph{implicit memory}, we reproduce the SSM-based
memory module of VideoSSM~\citep{yu2025videossm}. The four re-trained
baselines share the same Wan2.1 backbone, the same memory-token
budget, and the same training data and schedule as our model, so that
comparisons isolate the effect of the memory design rather than
backbone capacity or compute.

\paragraph{Implementation Details.}
We use Wan2.1~\citep{wan2025wan21} as the diffusion backbone. Actions
are injected by adding the action embedding to the timestep embedding,
following MIND-World~\citep{ye2026mind}. The memory size $N_m$ is set
to match the patch-token count of $20$ latent frames. We set the
compact stride in the memory encoder to $s=2$, giving a $4\times$
reduction in attention tokens, and set the pruning budget to $K=200$,
as defined in Sec.~\ref{sec:method_prune}. The memory encoder, the
geometry head, and the backbone are trained jointly under
$\mathcal{L}_{FM} + \lambda\,\mathcal{L}_{geo}$ with $\lambda = 0.05$,
following Sec.~\ref{sec:method_train}. We optimize for $8$K steps at
learning rate $1\mathrm{e}{-5}$ with batch size $32$.

\paragraph{Metrics.}
We evaluate three groups of properties. (1) \emph{Memory consistency}
is measured per frame between the rollout and the ground-truth video
with MSE, PSNR, SSIM, and LPIPS, following MIND's long-context memory
protocol~\citep{ye2026mind}. (2) \emph{Action accuracy} reports
relative pose error in translation and rotation: we recover the
camera trajectory from each generated rollout with
ViPE~\citep{huang2025vipe}, align it to the ground truth via Sim(3)
Umeyama, and compute RPE following MIND. (3) \emph{Geometric
consistency} is measured by 3D reprojection: we estimate per-frame
depth and camera pose with Depth Anything
3~\citep{lin2025depthanything3}, compute cross-frame cycle
reprojection error, and normalize it to a $0$--$100$ score where
higher is better. Several entries are reported as ``--'' in
Table~\ref{tab:mind_main_results}
because we copy Matrix-Game 2.0 and MIND-World numbers directly from
the MIND benchmark, and no directly comparable public results are
available for the additional metrics we include.

\subsection{Comparison with State-of-the-Art Methods}
\label{sec:exp_main}

\paragraph{Quantitative Results.}
\label{sec:exp_quantitative}

Table~\ref{tab:mind_main_results}
reports quantitative comparisons on the first-person and third-person
MIND subsets. GIM-World achieves the best memory-consistency scores
on both views, improving all four reconstruction metrics over explicit
and implicit memory baselines. The gain is especially clear on 3D
geometric consistency: our normalized reprojection score reaches
$81.70$ in the first-person setting and $87.10$ in the third-person
setting, compared with $73.97$ and $70.82$ for the strongest baseline
with comparable metrics.

The comparison across memory types supports the motivation of our
design. Explicit-memory methods keep historical observations available
to the generator, but retaining frames or packed tokens does not by
itself produce a compact cross-view world state, and their reprojection
scores remain substantially lower. The SSM baseline provides a compact
implicit state, but without geometry supervision it performs worse on
both reprojection and memory consistency. In contrast, GIM-World keeps
the fixed-size advantage of implicit memory while supervising that
state to encode view-consistent geometry. Action accuracy is also at
least on par with the strongest baseline on every subset, indicating
that improving the memory's geometric structure does not come at the
cost of controllable rollout dynamics.

\paragraph{Qualitative Results.}
\label{sec:exp_qualitative}

Figures~\ref{fig:qual_1st} and~\ref{fig:qual_3rd} compare GIM-World
against FramePack, CaM, and SSM on first-person and third-person
MIND scenes, with each row a
method and each column a rollout frame at a fixed stride over $400$
frames; all methods share the same initial observation and action
sequence. Baselines show clear structural collapse: scene layouts
drift within the first hundred frames, walls and landmark structures
deform or disappear, and the predicted content soon stops
corresponding to the same environment as the input. The third-person
setting additionally exposes character behavior, where baselines
either lose track of the controllable character or render it in
inconsistent locations. GIM-World shows only small viewpoint offsets
on individual frames, while the overall scene geometry, layout,
landmark structures, and character pose stay consistent with the
ground truth across the entire horizon.

Figure~\ref{fig:long_horizon} pushes the rollout to a thousand frames
on a single first-person scene from our model alone, sampled every
$50$ steps. The yellow boxes at frame~$100$ and frame~$750$ mark two
close viewpoints that the camera revisits more than six hundred
frames apart, and within these frames the small objects highlighted
by the green and red boxes retain the same spatial positions,
relative arrangement, and fine texture details across this long gap.
This indicates that the implicit memory stores a stable $3$D
representation of the scene rather than merely propagating short-term
context, and the world geometry remains globally consistent over a
thousand-frame autoregressive rollout.

\subsection{Ablation Study}
\label{sec:exp_ablation}

\begin{figure*}[t]
  \centering
  \includegraphics[width=\linewidth]{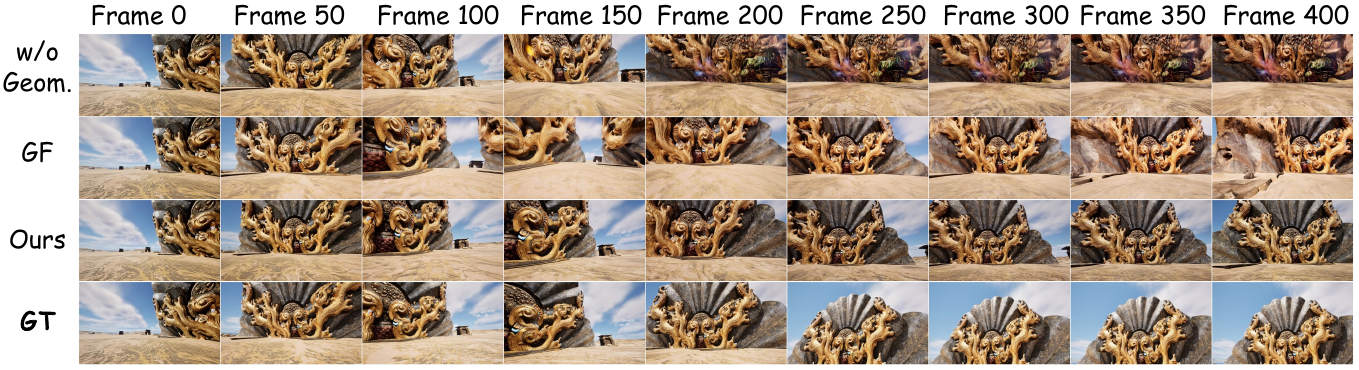}
  \caption{Qualitative ablation on geometry supervision on the MIND benchmark.}
  \label{fig:geo_qual}
\end{figure*}

\paragraph{Geometry Supervision.}
Table~\ref{tab:ablation_geometry} compares three ways of training the
memory encoder. \emph{w/o geometry} drops the geometry loss and trains
only with flow matching. \emph{Geometry Forcing} reproduces
\citet{wu2025geometry} by distilling VGGT~\citep{wang2025vggt}
features into intermediate diffusion layers, applying the geometry
signal on the generator rather than the memory. \emph{Ours} is the
camera-queryable supervision of Sec.~\ref{sec:method_geo}, applied
directly on the memory state.

Geometry Forcing improves only marginally over the no-geometry
baseline, lifting the reprojection score from $58.80$ to $63.40$,
while our supervision pushes it to $81.70$, with a parallel drop in
rotation RPE from $0.4823$ to $0.4126$. MSE and translation RPE move
much less because they are not primarily geometry-bound. Together with
Fig.~\ref{fig:geo_qual}, where Geometry Forcing still deviates in
scale and viewpoint while ours keeps the rollout aligned with the
ground-truth structure throughout the horizon, this confirms that
constraining the memory directly is more effective than constraining
the generator alone.

\begin{table}[t]
  \centering
  \small
  \setlength{\tabcolsep}{5pt}
  \setlength{\abovecaptionskip}{2pt}
  \setlength{\belowcaptionskip}{2pt}
  \renewcommand{\arraystretch}{1.0}
  \caption{
    Ablation on geometry supervision on the first-person MIND test
    set. The shaded row marks our setting.
  }
  \label{tab:ablation_geometry}
  \resizebox{\columnwidth}{!}{%
  \begin{tabular}{lccccc}
    \toprule
    Variant
    & MSE$\downarrow$
    & LPIPS$\downarrow$
    & Trans.$\downarrow$
    & Rot.$\downarrow$
    & Reproj.$\uparrow$ \\
    \midrule
    w/o geometry
    & 0.0628 & 0.6485 & 0.0252 & 0.5028 & 58.80 \\
    Geometry Forcing
    & \underline{0.0623} & \underline{0.6452} & \underline{0.0251}
    & \underline{0.4823} & \underline{63.40} \\
    \rowcolor{selectedrow}
    Ours
    & \textbf{0.0614} & \textbf{0.6304} & \textbf{0.0247}
    & \textbf{0.4126} & \textbf{81.70} \\
    \bottomrule
  \end{tabular}%
  }
\end{table}

\paragraph{Information-Guided Pruning.}
Table~\ref{tab:ablation_pruning} compares three pruning rules under
the same budget $K=200$: \emph{Uniform} temporal sampling,
\emph{Camera FPS} that performs farthest-point sampling in a
pose-time space, and our \emph{MI greedy} from
Sec.~\ref{sec:method_prune}. Uniform and Camera FPS sit close
together, splitting wins across metrics, while MI greedy is strictly
the best on every metric with a clear margin, showing that the
GP-based information criterion captures structure across pose,
viewing direction, and time that pure temporal or pose-space
heuristics miss.

\begin{table}[t]
  \centering
  \small
  \setlength{\tabcolsep}{7pt}
  \setlength{\abovecaptionskip}{2pt}
  \setlength{\belowcaptionskip}{2pt}
  \renewcommand{\arraystretch}{1.0}
  \caption{
    Ablation on history pruning on the first-person MIND test set. The
    shaded row marks our setting.
  }
  \label{tab:ablation_pruning}
  \resizebox{\columnwidth}{!}{%
  \begin{tabular}{lccccc}
    \toprule
    Pruning
    & MSE$\downarrow$
    & LPIPS$\downarrow$
    & Trans.$\downarrow$
    & Rot.$\downarrow$
    & Reproj.$\uparrow$ \\
    \midrule
    Uniform
    & \underline{0.0658} & 0.6582 & 0.0269
    & \underline{0.4475} & 77.05 \\
    Camera FPS
    & 0.0664 & \underline{0.6553} & \underline{0.0258}
    & 0.4598 & \underline{77.52} \\
    \rowcolor{selectedrow}
    MI greedy
    & \textbf{0.0614} & \textbf{0.6304} & \textbf{0.0247}
    & \textbf{0.4126} & \textbf{81.70} \\
    \bottomrule
  \end{tabular}%
  }
\end{table}

\paragraph{Compact Attention.}
Table~\ref{tab:ablation_compact} compares four attention designs in
the memory encoder. \emph{Full attention} runs at the native token
resolution. \emph{Global Compact} $s=2$ applies $2{\times}2$ pooling
to both the attention and the FFN. \emph{Attention Compact} $s=2$ is
our default from Sec.~\ref{sec:method_arch}, pooling only the
attention branch and keeping the FFN at full resolution.
\emph{Attention Compact} $s=4$ uses the same recipe with stronger
pooling.

Full attention defines the quality upper bound but is more than
$10\times$ slower than the compact variants. Our default matches it
on MSE and Reproj and is in fact the best on LPIPS, while costing
only $0.26\%$ of the DiT decode. Pooling the FFN as well in Global
Compact loses heavily on every metric at almost no extra speedup,
indicating that the FFN capacity is what the memory cannot afford to
lose. Stronger pooling at $s=4$ is cheaper still but degrades both
LPIPS and Reproj, so $s=2$ is the practical default.

\begin{table}[t]
  \centering
  \small
  \setlength{\tabcolsep}{4pt}
  \setlength{\abovecaptionskip}{2pt}
  \setlength{\belowcaptionskip}{2pt}
  \renewcommand{\arraystretch}{1.0}
  \caption{
    Ablation on compact attention. Runtime is measured relative to the
    DiT backbone. The shaded row marks the default setting.
  }
  \label{tab:ablation_compact}
  \resizebox{\columnwidth}{!}{%
  \begin{tabular}{lcccc}
    \toprule
    Variant
    & Time / DiT$\downarrow$
    & MSE$\downarrow$
    & LPIPS$\downarrow$
    & Reproj.$\uparrow$ \\
    \midrule
    Full attention
    & 2.80\% & \textbf{0.0608} & 0.6512 & \textbf{81.85} \\
    Global Compact ($s=2$)
    & \underline{0.20\%} & 0.0696 & 0.6892 & 62.45 \\
    \rowcolor{selectedrow}
    Attention Compact ($s=2$)
    & 0.26\% & \underline{0.0614} & \textbf{0.6304} & \underline{81.70} \\
    Attention Compact ($s=4$)
    & \textbf{0.10\%} & 0.0651 & \underline{0.6498} & 76.18 \\
    \bottomrule
  \end{tabular}%
  }
\end{table}

\section{Conclusion}

We presented \textbf{GIM-World}, a geometry-aware implicit memory framework
for long-horizon video world models. It compresses variable-length history
into fixed-size memory tokens, distills camera-queryable geometry into the
memory during training, and prunes redundant history to keep encoding bounded.
The geometry teacher is discarded at inference, leaving a lightweight memory
module. On MIND, GIM-World better preserves scene geometry and appearance
consistency than existing baselines.

{
    \small
    \bibliographystyle{ieeenat_fullname}
    \bibliography{references}
}

\end{document}